\documentclass[letterpaper, 10 pt, journal, twoside]{IEEEtran}
\usepackage[left=1.88cm,right=1.88cm,top=1.88cm,bottom=1.88cm]{geometry}
\usepackage{amsmath,amssymb,amsfonts}
\usepackage{algorithmic}
\usepackage{graphicx}
\usepackage{textcomp}
\usepackage{xcolor}
\usepackage{bbm}
\usepackage{tabularx, booktabs}
\usepackage[style=ieee, maxcitenames=2, mincitenames=1, natbib=true, dashed=false]{biblatex}
\usepackage{soul,color}
\usepackage{siunitx}
\usepackage{dsfont}
\usepackage{colortbl}
\usepackage{xcolor}

\usepackage{bbding}
\usepackage{amsmath}
\usepackage{graphicx} 
\usepackage{subcaption} 
\usepackage{xspace}
\usepackage{multirow}

\begin{document}

\title{
    CMP: Cooperative Motion Prediction with \\Multi-Agent Communication

    \author{Zehao Wang$^{1*}$, Yuping Wang$^{2*}$, Zhuoyuan Wu$^{3*}$, Hengbo Ma$^{4}$, Zhaowei Li$^{5}$, Hang Qiu$^{1\ddagger}$, and Jiachen Li$^{1\ddagger}$

    \thanks{Manuscript received: September 26, 2024; Revised: January 1, 2025; Accepted: February 4, 2025. This paper was recommended for publication by Editor M. Ani Hsieh upon evaluation of the Associate Editor and Reviewers' comments. $^*$ Equal contribution in random order \ $^\ddagger$ Corresponding authors}
    \thanks{$^{1}$Z. Wang, H. Qiu, and J. Li are with the University of California, Riverside, CA, USA. {\tt\footnotesize \{zehao.wang1, hangq, jiachen.li\}@ucr.edu}}%
    \thanks{$^{2}$Y. Wang is with the University of Michigan, Ann Arbor, MI, USA.}%
    \thanks{$^{3}$Z. Wu is an independent researcher.}%
    \thanks{$^{4}$H. Ma is with the University of California, Berkeley, CA, USA.}%
    \thanks{$^{5}$Z. Li is with the University of Washington, WA, USA.}%
    \thanks{Digital Object Identifier (DOI): 10.1109/LRA.2025.3546862.}
    }
}

\markboth{IEEE Robotics and Automation Letters. Preprint Version. Accepted February, 2025}
{Wang \MakeLowercase{\textit{et al.}}: CMP: Cooperative Motion Prediction with Multi-Agent Communication} 

\maketitle

\begin{abstract}
The confluence of the advancement of Autonomous Vehicles (AVs) and the maturity of Vehicle-to-Everything (V2X) communication has enabled the capability of cooperative connected and automated vehicles (CAVs). 
Building on top of cooperative perception, this paper explores the feasibility and effectiveness of cooperative motion prediction. 
Our method, CMP, takes LiDAR signals as model input to enhance tracking and prediction capabilities. 
Unlike previous work that focuses separately on either cooperative perception or motion prediction, our framework, to the best of our knowledge, is the first to address the unified problem where CAVs share information in both perception and prediction modules.
Incorporated into our design is the unique capability to tolerate realistic V2X transmission delays, while dealing with bulky perception representations.
We also propose a prediction aggregation module, which unifies the predictions obtained by different CAVs and generates the final prediction.
Through extensive experiments and ablation studies on the OPV2V and V2V4Real datasets, we demonstrate the effectiveness of our method in cooperative perception, tracking, and motion prediction. 
In particular, CMP reduces the average prediction error by 12.3\% compared with the strongest baseline.
Our work marks a significant step forward in the cooperative capabilities of CAVs, showcasing enhanced performance in complex scenarios. More details can be found on the project website: \url{https://cmp-cooperative-prediction.github.io}.
\end{abstract}


\section{Introduction}\label{sec:intro}

\IEEEPARstart{T}{he} current autonomous driving system is critically dependent on its onboard perception. Similar to human drivers, however, such dependency is vulnerable to situations with occlusions or impaired visibility. 
Leveraging multiple vantage points, cooperative perception~\cite{AVR,autocast, wang2020v2vnet, opv2v, xu2022cobevt} uses Vehicle-to-Everything (V2X) communications to share sensory information among connected and automated vehicles (CAVs) and infrastructure. This shared information varies in format, including raw data, processed features, or detected objects. Fusing this information from multiple viewpoints to the perspective of a vehicle recipient, the \textit{augmented} onboard perception can now ``see" beyond their direct line of sight and through occlusions. 

\begin{figure}[!]
    \centering
    \includegraphics[width=\columnwidth]{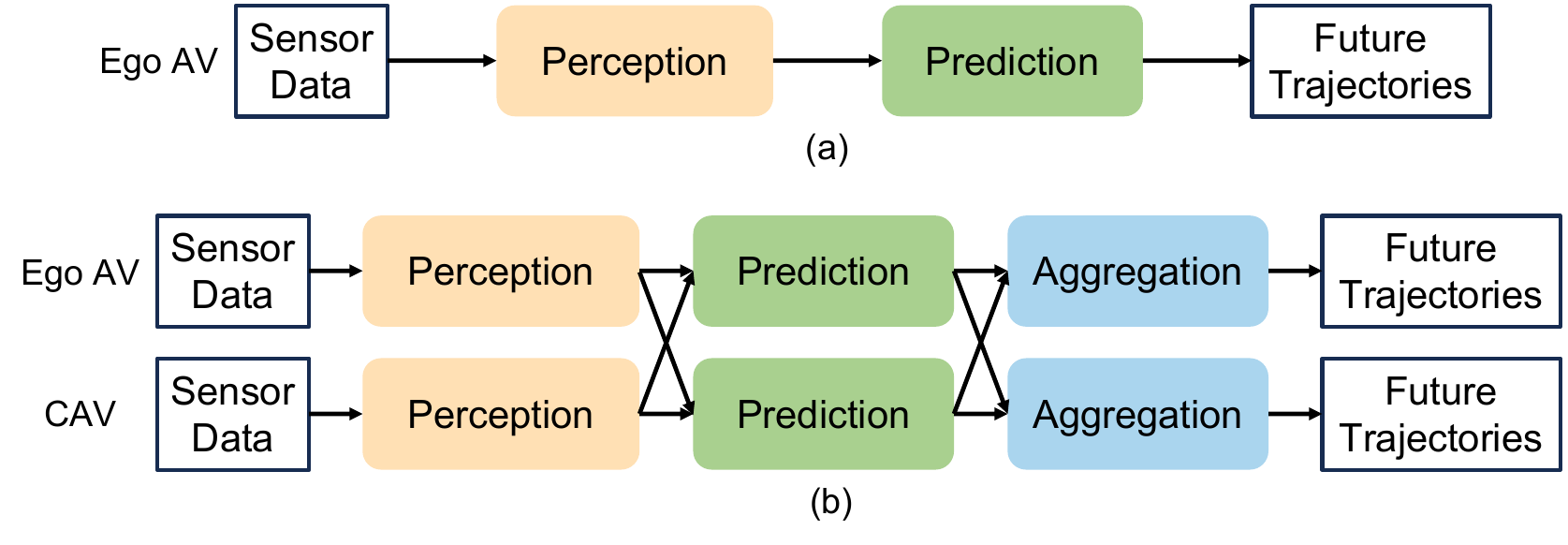}
    \vspace{-0.4cm}
    \caption{A comparison between the traditional pipeline and the proposed multi-vehicle cooperative prediction pipeline. (a) The traditional pipeline conducts perception and prediction based on a single AV's raw sensor data. (b) The proposed pipeline involves multiple cooperative CAVs, which share information to enhance both perception and prediction.}
    \vspace{-0.5cm}
    \label{fig:teaser}
\end{figure}

Current V2V research has largely been confined to either cooperative perception or motion prediction, with no comprehensive studies on their joint application. 
Beyond object detection, most works incorporate other tasks, such as prediction~\cite{wang2020v2vnet} and mapping~\cite{xu2022v2xvit} as auxiliary outputs. \citet{wang2020v2vnet} proposes a V2V method for perception and prediction, which transmits intermediate representations of point cloud features.
However, integrating perception and prediction, as illustrated in Fig. \ref{fig:teaser}(b), to fully realize V2V cooperation remains unexplored.
On motion prediction, initial efforts~\cite{hu2020collaborative, Choi2021prediction, v2voffloading} use LSTM-based networks on simple datasets. Recent studies~\cite{shi2023motion, wang2023eqdrive} adopt attention networks and graph convolutional networks to enhance motion prediction. 
However, these approaches rely on the ground-truth trajectory data, neglecting the uncertainties and inaccuracies propagated from upstream detection and tracking tasks. This reliance on ground truth data remains insufficient to address the real-world challenge of handling uncertain trajectories, underscoring the need for research that integrates perception and prediction in V2V cooperation.

To fill the gap between cooperative perception and motion prediction, we introduce a novel framework for cooperative motion prediction based on the raw sensor data.
To the best of our knowledge, we are the first to develop a practical method that jointly solves the perception and prediction problem with CAV communications in both components.
Each CAV computes its own bird-eye-view (BEV) feature representation from its LiDAR point cloud. The data is processed, compressed, and broadcast to nearby CAVs. 
The ego CAV fuses the received features and performs detections on surrounding agents. 
A multi-object tracker then generates the historical trajectories of the surrounding objects. Each CAV then performs motion prediction. After that, the individually predicted trajectories from each connected vehicle are broadcast again.
As our model collects the predictions from surrounding CAVs, the predictions and intermediate features from perception are used to refine the motion predictions. Our method allows for realistic transmission delays between CAVs and bandwidth limitations. 

In this paper, our main contributions are as follows:
\begin{itemize}
  \item We propose a practical, latency-robust framework for cooperative motion prediction, which leverages the information shared by multiple CAVs to enhance perception and motion prediction performance.
  \item We develop an attention-based prediction aggregation module to take advantage of the predictions shared by other CAVs, which improves prediction accuracy. 
  \item Our method achieves state-of-the-art performance in cooperative prediction under practical settings on the OPV2V and V2V4Real datasets.
\end{itemize}

\section{Related Work}\label{sec:relatedwork}

\subsection{Cooperative Perception}
Cooperative perception enhances field-of-view by sharing data among CAVs. Previous works have developed early fusion techniques based on shared raw LiDAR or RGB camera data~\cite{autocast}. However, it requires high transmission bandwidth. Another strategy, late fusion, allows vehicles to only share their final detections ~\cite{latefusion}. However, in real-life deployments, the performance of late fusion is capped by the loss of context information and individual detection accuracy. 
To balance this trade-off, the middle-ground strategy of intermediate fusion~\cite{coopernaut, wang2020v2vnet, qiao2023adaptive, xu2022cobevt, xiang2023hmvit} has become more prevalent. In this strategy, CAVs encode and share intermediate features for fusion. 
For example, V2VNet~\cite{wang2020v2vnet} employed a Graph Neural Network to aggregate information from different viewpoints. 
AttFuse~\cite{opv2v} deployed an attention mechanism to fuse the intermediate features. \citet{qiao2023adaptive} proposed a fusion model that adaptively chooses intermediate features for better integration.

\vspace{-0.3cm}
\subsection{Motion Prediction}
Motion prediction in single AVs has advanced through graph-based and transformer-based models~\cite{li2020evolvegraph,gao2020vectornet,toyungyernsub2022dynamics,li2021spatio,varadarajan2021multipath++,choi2021shared,sun2022m2i,lange2024scene,shi2023motion, girase2021loki, dax2023disentangled, ruan2023learning, li2023game, hivt, sun2022m2i, gao2020vectornet, wang2023equivariant}.
Recent work introduces a transformer model into their models due to its efficient attention mechanism and ability to handle long-term prediction. MTR~\cite{shi2023motion} uses motion query pairs where each pair is in charge of one motion mode prediction, which is more efficient than goal-based strategies~\cite{gu2021densetnt} and converges faster than direct regression strategies~\cite{varadarajan2021multipath++, ngiam2022scene}.

\vspace{-0.3cm}
\section{Problem Formulation}
The goal of the cooperative prediction task is to infer the future trajectories of all the movable agents in the scene that can be detected by multiple collaborative CAVs with onboard sensors. In this work, we only use the LiDAR information for perception (i.e., object detection and tracking) to obtain the agents' trajectories.
We denote the number of CAVs as $N_{\text{CAV}}$, the LiDAR point cloud of $i$-th CAV at time $t$ as $\mathbf{L}^i_{t}, i = 1,...,N_{\text{CAV}}$, and the local map information as $\mathbf{M}^i_{t}$.
Assume that there are $N_t$ detected agents at time $t$, we denote their historical trajectories as $\mathbf{X}_{t-T_\text{h}+1:t}$ where $T_\text{h}$ represents the history horizon.
We aim to infer their multi-modal future trajectories $\hat{\mathbf{X}}_{t+1:t+T_\text{f}}$ based on the above information where $T_\text{f}$ represents the prediction horizon.

\section{Method}\label{sec:methods}

\begin{figure*}
    \centering
    \includegraphics[width=0.95\textwidth]{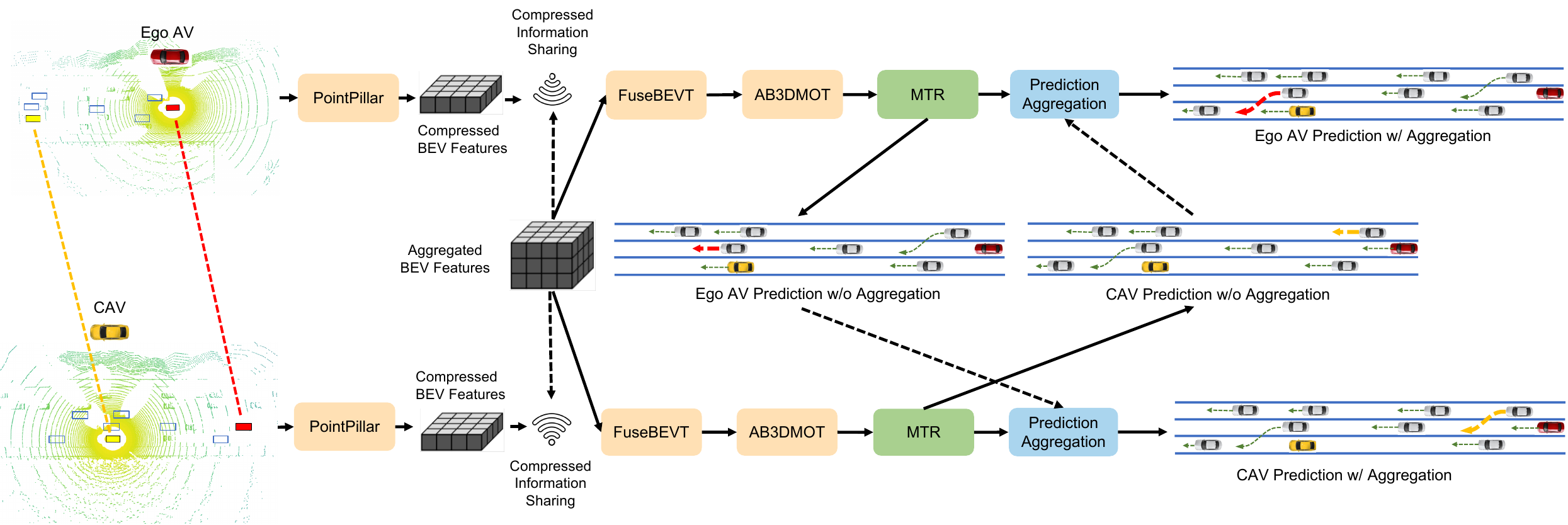}
    \caption{An overall diagram of the proposed cooperative motion prediction pipeline.}
    \vspace{-0.6cm}
    \label{fig:method}
\end{figure*}

\subsection{Method Overview}

Fig. \ref{fig:method} provides an overall diagram of our CMP framework, which consists of three major components: cooperative perception, trajectory prediction, and prediction aggregation. 
The cooperative perception module takes in the raw sensor data obtained by CAVs and generates the observed agents' trajectories through object detection and multi-object tracking. 
The trajectory prediction module then takes in the historical observations and infers future trajectories from the perspective of each CAV. Finally, the prediction aggregation module leverages the predictions from all CAVs and generates the final prediction hypotheses.

\vspace{-0.3cm}
\subsection{Cooperative Perception}
\label{perception}
The cooperative perception module aims to detect and track objects based on the 3D LiDAR point clouds obtained by multiple CAVs. We modify CoBEVT \cite{xu2022cobevt} as the backbone of the object detection model followed by the AB3DMOT tracker \cite{DBLP:conf/iros/WengWHK20} to obtain historical trajectories of agents.

\textbf{Cooperative Object Detection.}
PointPillar~\cite{Lang_Vora_Caesar_Zhou_Yang_Beijbom_2019} is employed to extract point cloud features for each CAV with a voxel resolution of (0.4, 0.4, 4) along $x$, $y$, and $z$ axes.  Each CAV $i$ calculates an unified bird-eye-view (BEV) feature $\mathbf{F}^i \in \mathbb{R}^{H \times W \times C}$, where $H$, $W$, and $C$ denote height, width, and channels, respectively.

Due to the real-world hardware constraints on the volume of the transmitted data for V2V applications, it is necessary to compress the BEV features before transmission to avoid large bandwidth-induced delays. 
As in \cite{xu2022cobevt}, a convolutional auto-encoder is used for feature compression and decompression. 
Upon receipt of the broadcast messages containing intermediate BEV representations and the sender's pose, a differentiable spatial transformation operator $\mathbf{\Gamma_{\xi}}$ is used to align the features to the ego vehicle's coordinate, which is written as $\mathbf{H}^i=\mathbf{\Gamma_{\xi}}\left(\mathbf{F}^i\right) \in \mathbb{R}^{H \times W \times C}$.
The operator learns spatial transformations on the input to enhance the geometric invariance of the model, addressing localization errors present in real-world scenarios \cite{xu2023v2v4real}.
We assume that CAVs' onboard clocks are synchronized via GPS/GNSS signals~\cite{wang2020v2vnet} and can process shared features in a \SI{100}{ms}-synchronized fashion. This means that receivers will wait and process the shared information in the fixed 100 ms latency. Frames taking longer than \SI{100}{ms} to transmit will be dropped. When there is a range of random delays within \SI{100}{ms}, the receivers are assumed to process data delayed by one frame. The above assumptions are consistent with cooperative perception literature~\cite{xu2022cobevt, xu2022v2xvit, opv2v, autocast, AVR}.
Then, FuseBEVT \cite{xu2022cobevt} is used to merge the BEV features received from various agents. 
More specifically, the ego vehicle first aggregates all the available features into a tensor $\mathbf{h}\in \mathbb{R}^{N_\text{CAV} \times H\times W\times C}$, which is then processed by the FuseBEVT module to obtain the fused feature $\mathbf{h'}\in \mathbb{R}^{H\times W\times C}$. 
Finally, two $3 \times3$ convolutional layers are applied for classification and regression to obtain the 3D bounding boxes of objects.
CoBEVT outputs a collection of detections at time $t$ denoted by $\mathbf{D}_{t}=\{\mathbf{D}^1_t,..., \mathbf{D}_t^{N_t}\}$, where $N_t$ represents the total number of detections. Each detection $D_t^j$ is characterized by a tuple $(x, y, z, \theta, l, w, h, s)$, which encapsulates the 3D coordinates of the object's center ($x$, $y$, $z$), the 3D dimensions of the object bounding box ($l$, $w$, $h$), the orientation angle $\theta$, and the confidence score $s$.

\textbf{Multi-Object Tracking.}
The tracking module associates the detected 3D bounding boxes of objects into trajectory segments. We adopt AB3DMOT \cite{DBLP:conf/iros/WengWHK20}, an online multi-object tracking algorithm, which takes in the detections in the current frame and the associated trajectories in previous frames. 
Excluding the pre-trained cooperative object detection module, AB3DMOT requires no additional training and is simply applicable for inference.
More specifically, after obtaining the 3D bounding boxes from the cooperative object detection module, we apply a 3D Kalman filter to predict the state of the associated trajectories from previous frames to the current frame. Then, a data association module is adopted to match the predicted trajectories from the Kalman filter and the detected bounding boxes in the current frame. The 3D Kalman filter updates the state of matched trajectories based on the matched detections. Throughout the tracking process, a birth and death memory creates trajectories for new objects and deletes trajectories for disappeared objects. More details of these operations can be found in \cite{DBLP:conf/iros/WengWHK20}.
The tracker outputs the historical trajectories of all the agents detected at time $t$, denoted as $\mathbf{X}_{t-T_\text{h}+1:t}$, which serves as the input of the trajectory prediction module.

\vspace{-0.3cm}
\subsection{Motion Prediction}
\label{trajectory_prediction}
Our trajectory prediction module is built upon MTR~\cite{shi2023motion}, a state-of-the-art model consisting of a scene context encoder and a motion decoder. 
We only provide a general introduction, and more details about the model can be found in \cite{shi2023motion}.

For the $i$-th CAV, the scene context encoder extracts features from the agents' trajectories $\mathbf{X}_{t-T_\text{h}+1:t}$ and the local map information $\mathbf{M}^i_{t}$. The agents' trajectories are represented as polyline vectors \cite{gao2020vectornet}, which are processed by a PointNet-like polyline encoder \cite{qi2017pointnet} to extract agent features.
The map information is encoded by a Vision Transformer \cite{dosovitskiy2020image} to extract map features.
Then, a Transformer encoder is used to capture the local scene context. Each layer uses multi-head attention with queries, keys, and values defined relative to previous layer outputs and position encodings, integrating the trajectory embeddings and map embeddings. 
Future agent movements are predicted via regression based on the extracted past agent features. 
These predictions are re-encoded by the same polyline encoder and merged with historical context features.

After obtaining the scene context features, a Transformer-based motion decoder is employed to generate multi-modal prediction hypotheses through joint optimization of global intention localization and local movement refinement.
More specifically, $K$ representative intention points $\mathbf{I}\in \mathbb{R}^{K\times 2}$ are generated by adopting the $k$-means clustering algorithm on the endpoints of ground truth trajectories ($K=64$ in our setting), where each intention point represents an implicit motion mode that represents the motion direction. 
The local movement refinement enhances global intention localization by iteratively refining trajectories with fine-grained trajectory features.
The dynamic searching query is initially set at the intention point, and updates dynamically based on the trajectory predicted at each decoder layer, serving as a spatial point's position embedding.

In the decoder, static intention queries transmit information across motion intentions while dynamic searching queries gather trajectory-specific information from the scene context. 
The updated motion query is expressed as $\mathbf{C}^j\in \mathbb{R}^{K\times D}$ in the $j$-th layer where $D$ is the feature dimension.
Each decoder layer adds a prediction head to $\mathbf{C}^j$ for creating future trajectories. Due to the multi-modal nature of agents' behaviors, a Gaussian Mixture Model (GMM) is adopted for trajectory distributions. 
For each future time step $t' \in \{t+1,...,t+T_\text{f}\}$, we infer the likelihood $p$ and parameters ($\mu_x, \mu_y, \sigma_x, \sigma_y, \rho$) of Gaussian components by 
\begin{equation}\label{eq:gau_reg}
    \mathbf{Z}^j_{t+1:t+T_\text{f}} = \text{MLP}(\mathbf{C}^j),
\end{equation}
where $\mathbf{Z}^j_{t'}\in \mathbb{R}^{{K}\times 6}$ contains the parameters of $K$ Gaussian components $\mathcal{N}_{1:\mathcal{K}}(\mu_x, \sigma_x; \mu_y, \sigma_y; \rho)$ and the corresponding likelihoods $p_{1:K}$. The distribution of the agent's position at time $t'$ is written as 
\begin{equation}\label{eq:gau_prob}
    P^j_{t'}(o) = \sum_{k=1}^{K}p_k\cdot\mathcal{N}_k(o_x-\mu_x, \sigma_x;o_y - \mu_y, \sigma_y;\rho),
\end{equation}
where $P^j_{t'}(o)$ denotes the probability of the agent located at $o\in \mathbb{R}^{2}$ at time $t'$. The trajectory predictions of all the agents $\hat{\mathbf{X}}_{t+1:t+T_\text{f}}$ can be derived from the center points of corresponding Gaussian components.

\vspace{-0.3cm}
\subsection{Prediction Aggregation}
Besides sharing the BEV features between CAVs, we also propose to transmit the prediction hypotheses generated by each CAV to others. Each CAV adopts an aggregation mechanism to fuse the predictions received from others with its own predictions.
The underlying intuition is that the predictions for a certain agent obtained from different CAVs may have different levels of reliability. For example, a CAV closest to the predicted agent may generate better predictions than others. Thus, the predictions from different CAVs may complement each other, leading to the best final prediction.
This mechanism is scalable and can effectively handle varying numbers of CAVs.
More specifically, in a scenario with $N_\text{CAV}$ CAVs and $N_o$ predicted agents, the GMM prediction components for agent $j$ by CAV $i$ at time $t$ are denoted as $\mathbf{Z}_{j, t+1:t+T_f}^i)$.
The local map and BEV features of CAV $i$ are denoted as $\mathbf{M}^i_{t}$ and $\mathbf{H}^i_t$, respectively.
Before applying the attention mechanism, the predictions from each CAV are concatenated with relevant contextual features, including local map information and BEV features. This concatenation enriches the input by providing spatial context, which is crucial for accurately aligning and integrating predictions from multiple CAVs.
We aggregate the GMM components of the predicted trajectories, BEV features, and map information for all CAVs.  
For CAV $i$, it begins the aggregation process by concatenating its GMM, map, and BEV features:
\begin{equation}
    \mathbf{E}_{j, t}^i = [\text{MLP}(f(\mathbf{Z}_{j, t+1:t+T_f}^i)), \text{MLP}(f(\mathbf{M}_t^i)), \text{MLP}(f(\mathbf{H}_t^i))],
\end{equation}
Upon receiving the GMM components from other CAVs $k$ ($1 \leq k \leq N_\text{CAV}, k \neq i$), the same map, BEV features from the ego are concatenated again:
\begin{equation}
    \mathbf{E}_{j, t-1}^k = [\text{MLP}(f(\mathbf{Z}_{j, t:t+T_f-1}^k)), \text{MLP}(f(\mathbf{M}_t^i)), \text{MLP}(f(\mathbf{H}_t^i))],
\end{equation}
followed by a multi-head self-attention to fuse the features across all CAVs,
\begin{equation}
    \mathbf{G}^{i}_{j, t} = \text{MHA}([\mathbf{E}_{j, t}^i,..., \mathbf{E}_{j, t-1}^{k}]), 1 \leq k \leq N_\text{CAV}, k \neq i
\end{equation}
where $\text{MHA}$ is multi-head self-attention, $f$ is the flatten operation, and $\mathbf{G}^{i}_{j, t}$ is the aggregated feature for agent $j$ from the perspective of CAV $i$. The GMM components from other CAVs are delayed by one frame.
The self-attention mechanism dynamically weighs predictions from multiple CAVs based on their contextual consistency with the BEV and map features. For scenarios where predictions are missing due to communication delays or occlusions, it adapts seamlessly by focusing on the available predictions.
Finally, two separate MLPs derive the aggregated Gaussian parameters by $\mathcal{N}^i_{j, 1:K, t+1:t+T_\text{f}}(\mu_x, \sigma_x; \mu_y, \sigma_y; \rho) = \ \text{MLP}(\mathbf{G}^{i}_{j, t})$ and $p^j_{j, 1:K,t+1:t+T_f} = \ \text{MLP}(\mathbf{G}^{i}_{j, t})$,
which will be used to sample the final prediction hypotheses.

\vspace{-0.3cm}
\subsection{Loss Functions}
\textbf{Cooperative Object Detection.}
We adopt the same loss function as CoBEVT~\cite{xu2022cobevt}. 
In particular, our framework incorporates the two convolutional layers for the detection head and employs the smooth $L1$ loss for bounding box localization $\mathcal{L}_\text{det\_loc}$ and the focal loss for classification $\mathcal{L}_\text{det\_cls}$, as outlined in~\cite{Lin_Goyal_Girshick_He_Dollar_2017}. The complete loss function is $\mathcal{L_{\text{det}}}=(\beta_\text{loc}\mathcal{L}_{\text{det\_loc}}+\beta_\text{cls}\mathcal{L}_{\text{det\_cls}}) / N_p$,
where $N_p$ denotes the count of positive instances, $\beta_\text{loc}=2/3$, and $\beta_\text{cls}=1/3$ .

\textbf{Motion Prediction.}
Our prediction model is trained with two loss terms. An $L1$ regression loss is used to refine the outputs in Eq.~\eqref{eq:gau_reg}. We also employ a negative log-likelihood loss based on Eq.~\eqref{eq:gau_prob} to enhance the prediction accuracy of the actual trajectories. 
We take the weighted average of these two terms as the total loss, which is written as
\begin{equation}\label{eq:prediction_loss}
\mathcal{L_{\text{pred}}}=\omega_\text{loc}\mathcal{L}_{\text{pred\_loc}}+\omega_\text{cls}\mathcal{L}_{\text{pred\_cls}}.
\end{equation}
Following \cite{varadarajan2021multipath++}, we apply a hard-assignment technique for optimization by choosing the motion query pair that is closest to the ground truth (GT) trajectory's endpoint as the positive Gaussian component, determined by the distance between each intention point and the GT endpoint. The Gaussian regression loss is applied at every decoder layer, and the overall loss combines the auxiliary regression loss with the Gaussian regression losses with equal weights.

\textbf{Prediction Aggregation.}
Our prediction aggregation module produces outputs in the same format as the motion prediction module, and we thus apply the same loss function as Eq.~\eqref{eq:prediction_loss} to supervise the final prediction.

\begin{table*}[!t]
\centering
\setlength{\tabcolsep}{2.pt}
\renewcommand{\arraystretch}{1.}
\caption{The comparisons of motion prediction performance under 100 ms delay and 256$\times$ compression rate (meter). Upper: OPV2V. Lower: V2V4Real.}
\label{tab:motion_prediction}
\label{prediction}
{
    \begin{tabular}{c|m{2.7cm}<{\centering}|m{2cm}<{\centering}|m{2cm}<{\centering}|m{2cm}<{\centering}|m{2cm}<{\centering}|m{2cm}<{\centering}|m{2cm}<{\centering}}
    \toprule
    \midrule
    Model & Cooperation Type & minADE$_6$@$1s\downarrow$ & minADE$_6$@$3s\downarrow$& minADE$_6$@$5s\downarrow$& minFDE$_6$@$1s\downarrow$& minFDE$_6$@$3s\downarrow$& minFDE$_6$@$5s\downarrow$ \\
    \midrule
    SinBEVT \cite{xu2022cobevt} & No Cooperation & 0.4099 & 1.1573 & 2.2217 & 0.6883 & 2.4583 & 5.1853 \\
    V2VNet \cite{wang2020v2vnet} & Cooperative P\&P & 0.4065 & 1.1127 & 2.1174 & 0.6452 & 2.3416 & 4.9037 \\
    CMP (Ours) & Cooperative Perception & 0.3543 & 1.0222 & 1.9616 & 0.5849 & 2.1772 & 4.5383 \\
    \textbf{CMP (Ours)} & Cooperative P\&P & \textbf{0.3429} & \textbf{0.9785} & \textbf{1.8578} & \textbf{0.5699} & \textbf{2.0243} & \textbf{4.1628} \\
    \bottomrule
    \end{tabular}
}

\vspace{0.3cm}
{
    \begin{tabular}{c|m{2.7cm}<{\centering}|m{2cm}<{\centering}|m{2cm}<{\centering}|m{2cm}<{\centering}|m{2cm}<{\centering}|m{2cm}<{\centering}|m{2cm}<{\centering}}
    \toprule
    \midrule
    Model & Cooperation Type & minADE$_6$@$1s\downarrow$ & minADE$_6$@$3s\downarrow$& minADE$_6$@$5s\downarrow$& minFDE$_6$@$1s\downarrow$& minFDE$_6$@$3s\downarrow$& minFDE$_6$@$5s\downarrow$ \\
    \midrule
    SinBEVT \cite{xu2022cobevt} & No Cooperation & 1.4469 & 3.4768 & 5.0250 & 2.6764 & 6.2796 & 8.6677 \\
    V2VNet \cite{wang2020v2vnet} & Cooperative P\&P & 1.4460 & 3.0980 & 4.4426 & 2.3143 & 5.5530 & 7.5711 \\
    CMP (Ours) & Cooperative Perception & 1.2077 & 2.8034 & 4.0857 & 2.1428 &  5.0322 & 7.3187 \\
    \textbf{CMP (Ours)} & Cooperative P\&P & \textbf{1.1398} & \textbf{2.6611} & \textbf{3.9099} & \textbf{2.0318} & \textbf{4.8249} & \textbf{7.0182} \\
    \bottomrule
    \end{tabular}
}
\vspace{-0.4cm}
\end{table*}

\section{Experiments}\label{sec:experiments}
\subsection{Dataset}
We use the OPV2V \cite{opv2v} and V2V4Real \cite{xu2023v2v4real} datasets to validate our approach. 
The OPV2V dataset contains 73 traffic scenarios with a duration of about 25 seconds with multiple CAVs. 
A range of two to seven CAVs may appear concurrently, which are equipped with a LiDAR sensor and four cameras from different views.
Following~\cite{xu2022cobevt}, we use a surrounding area of $\SI{100}{m} \times \SI{100}{m}$ with a map resolution of \SI{39}{cm} for evaluation. 
The V2V4Real dataset is a recent real-world multi-modal dataset for cooperative perception, which is collected by two vehicles equipped with LiDAR sensors and two mono cameras driving together through 67 scenarios in the USA. Each scenario lasts 10-20 seconds long.

\vspace{-0.3cm}
\subsection{Baselines and Evaluation Metrics}

To demonstrate the effectiveness of our method, we conduct ablation studies on various model components and compare our method with V2VNet \cite{wang2020v2vnet}, a state-of-the-art baseline that leverages V2V communication for joint perception and prediction. V2VNet does not have a tracking module, thus we do not include it in the comparison of tracking performance.
We employ the following evaluation metrics to compare our method with baselines:

\textbf{Motion Prediction.} We predict the agents' trajectories for the future 5.0 seconds based on 1.0 seconds of historical observations. We use the standard evaluation metrics as in \cite{shi2023motion}, including minADE$_6$ and minFDE$_6$.

\textbf{Object Detection.}
We use the standard evaluation metrics as~\cite{opv2v, DBLP:conf/icra/XuCXXLM23}, including Average Precision (AP), Average Recall (AR), and F1-score at the Intersection over Union (IoU) thresholds of 0.3, 0.5 and 0.7, respectively.

\textbf{Tracking.} We employ a set of standard evaluation metrics same as~\cite{DBLP:conf/iros/WengWHK20}, including Multi-Object Tracking Accuracy (MOTA), Average Multi-Object Tracking Accuracy (AMOTA), Average Multi-Object Tracking Precision (AMOTP), scaled Average Multi-Object Tracking Accuracy (sAMOTA), Mostly Tracked
Trajectories (MT), and Mostly Lost Trajectories (ML). 

\vspace{-0.3cm}
\subsection{Implementation Details}
\textbf{Object Detection.} 
CoBEVT \cite{xu2022cobevt} assumes no delay in the communication between CAVs, which may not be realistic due to hardware or wireless communication constraints.
To address this limitation, our model allows for up to a \SI{100}{ms} (i.e., 1 frame) delay in receiving the messages (i.e., BEV features) from other CAVs. In addition, our BEV features are compressed by 256 times compared with that in CoBEVT. 
Instead of selecting a single CAV as the ego vehicle in the original OPV2V and V2V4Real traffic scenarios as in \cite{xu2022cobevt} \cite{xu2023v2v4real}, we augment the training data samples by treating each of the CAVs in the scene as the ego vehicle. This leads to a more diverse and robust dataset, resulting in improved generalization and cooperative performance across various traffic scenarios. This also allows us to incorporate multiple CAVs fused BEV features in later prediction modules. We train our model using the AdamW optimizer with a learning rate scheduler starting at $1\times 10^{-3}$ and reduced every 10 epochs. 
We keep the consistent experiment setting for all baselines, including LiDAR range and a \SI{100}{ms} communication delay.

\textbf{Tracking}. 
In our setting, we set $F_\text{min}=3$ and $\text{Age}_\text{min}=2$ in the birth/death memory module. The data association module uses a threshold of $\text{IoU}_{\text{min}}=0.01$ for vehicles, and $\text{Dist}_{\text{max}}$ is set to 10. More details can be found in \cite{DBLP:conf/iros/WengWHK20}.

\textbf{Motion Prediction}. 
We use 6 encoder layers for context encoding with a hidden feature dimension of 256. 
The decoder employs 6 layers and 64 motion query pairs, determined by $k$-means clustering on the training set. 
We pre-train the prediction model with an AdamW optimizer with a learning rate of $1\times 10^{-4}$ and a batch size of 80 over 30 epochs.
More details can be found in \cite{shi2023motion}.

\textbf{Prediction Aggregation}.
We use three MLPs to encode the GMM parameters, map features, and BEV features, respectively. Then, an 8-head, 5-layer transformer encoder is used to aggregate the features, followed by two MLPs to decode the outputs into the final GMM trajectory and scores, which follow the same format as the outputs of the prediction module. 
We train the aggregation module with a learning rate of $1\times 10^{-4}$ and fine-tune the prediction module with a reduced learning rate of $1\times 10^{-6}$. The learning rates decay in the same manner as the prediction model. We train the model for 30 epochs with a batch size of 8.

\begin{figure*}[!]
    \centering
    \begin{subfigure}[b]{0.49\linewidth}
        \includegraphics[width=\linewidth,trim={0, 2cm, 0, 3cm},clip]{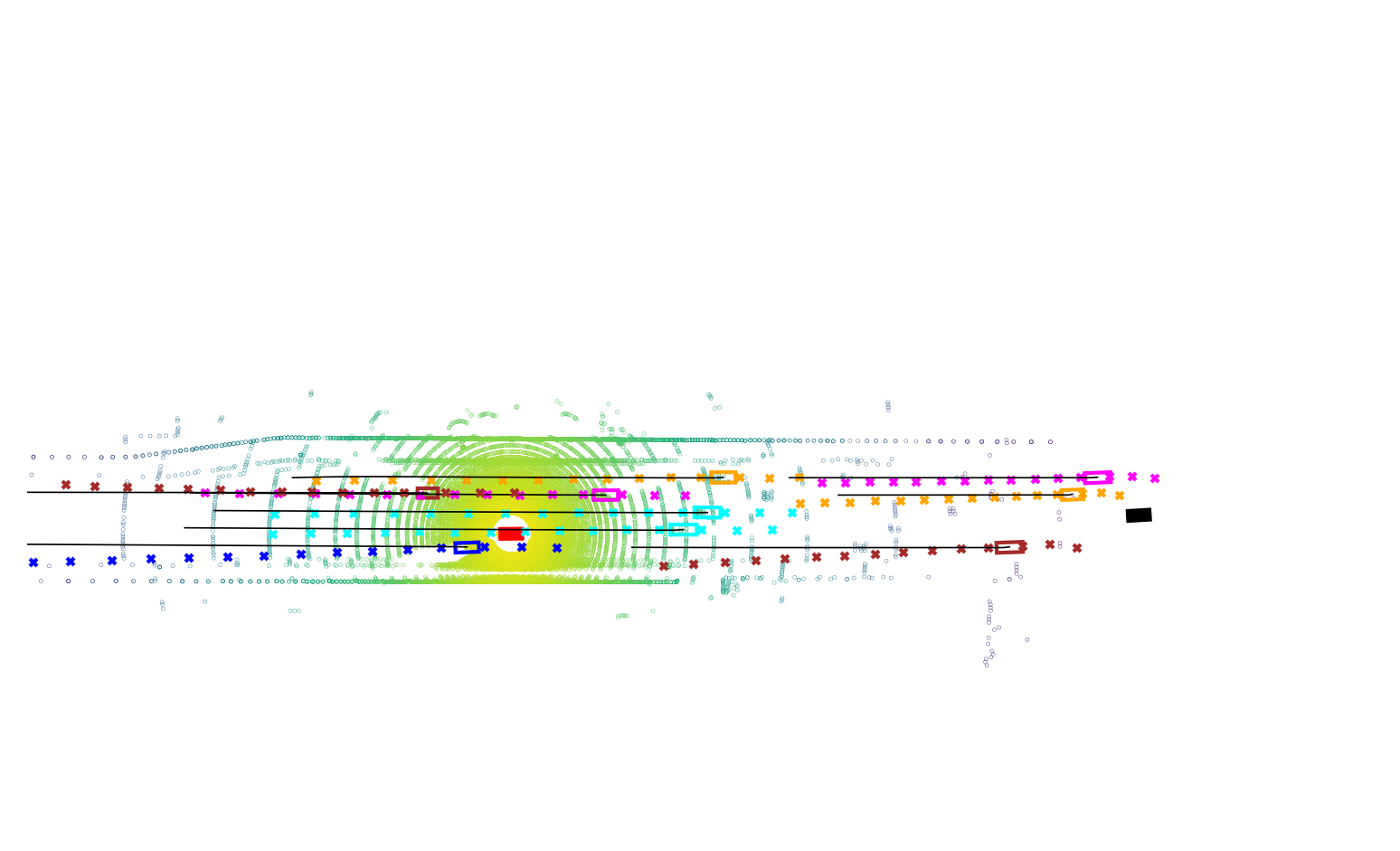}
        \caption{No Cooperation}
        \label{fig:case1n}
        
    \end{subfigure}
    \begin{subfigure}[b]{0.49\linewidth}
        \includegraphics[width=\linewidth,trim={0, 2cm, 0, 3cm},clip]{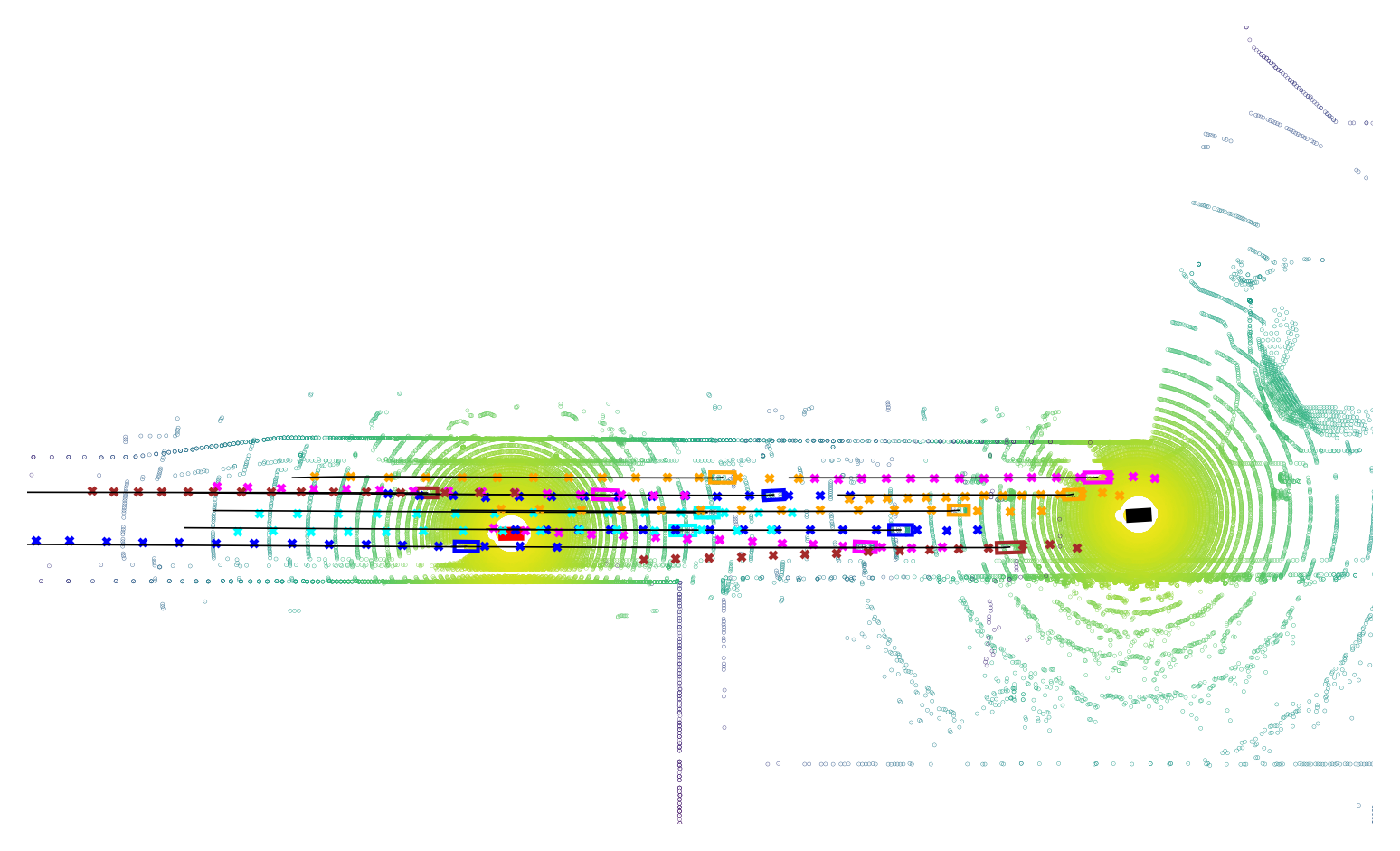}
        \caption{Cooperative Prediction (Ours)}
        \label{fig:case1y}
        
    \end{subfigure}
    
    \begin{subfigure}[b]{0.49\linewidth}
        \includegraphics[width=\linewidth,trim={0, 3cm, 0, 3cm},clip]{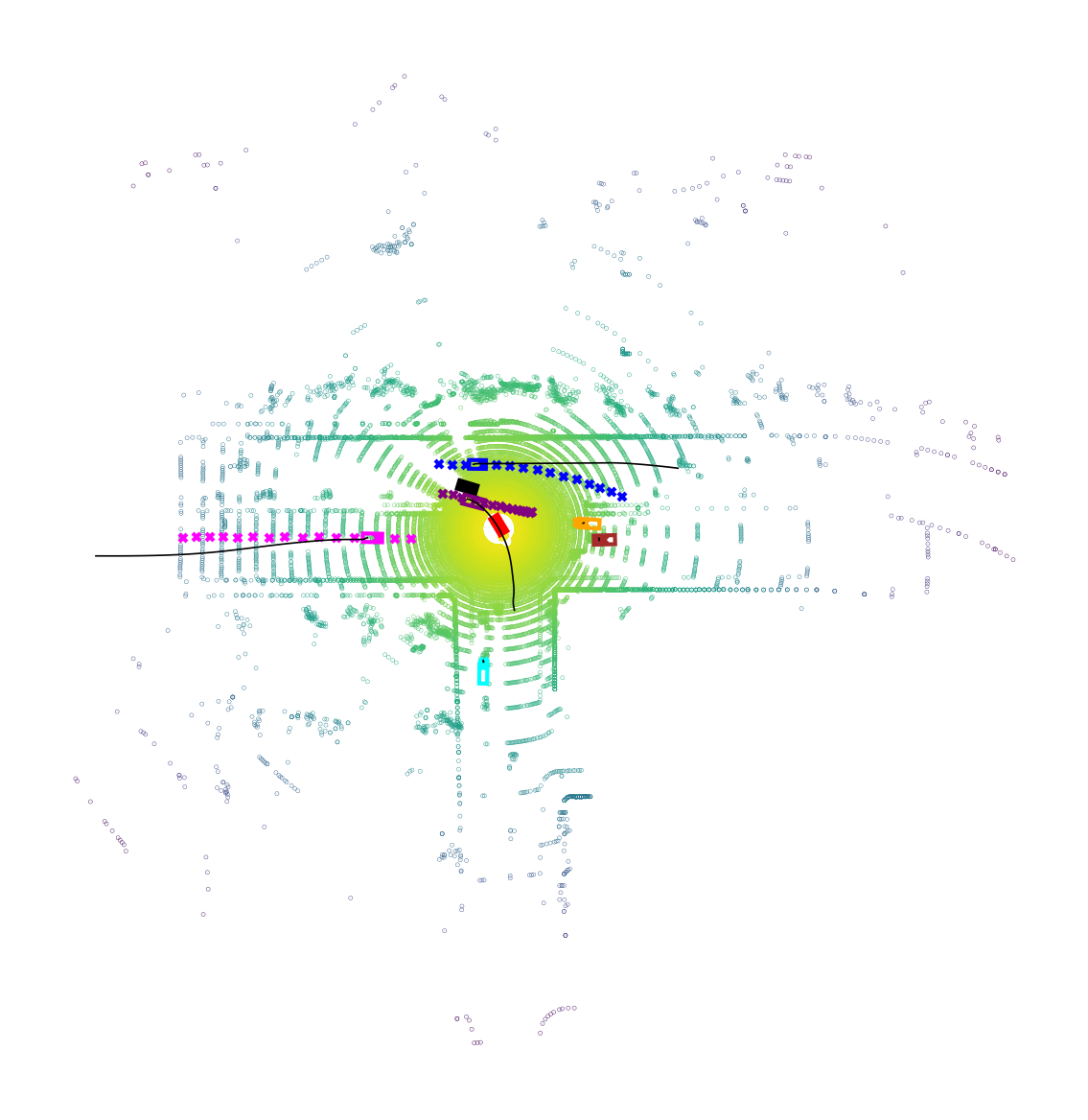}
        \caption{No Cooperation}
        \label{fig:case3n}
        
    \end{subfigure}
    \begin{subfigure}[b]{0.49\linewidth}
        \includegraphics[width=\linewidth,trim={0, 3cm, 0, 3cm},clip]{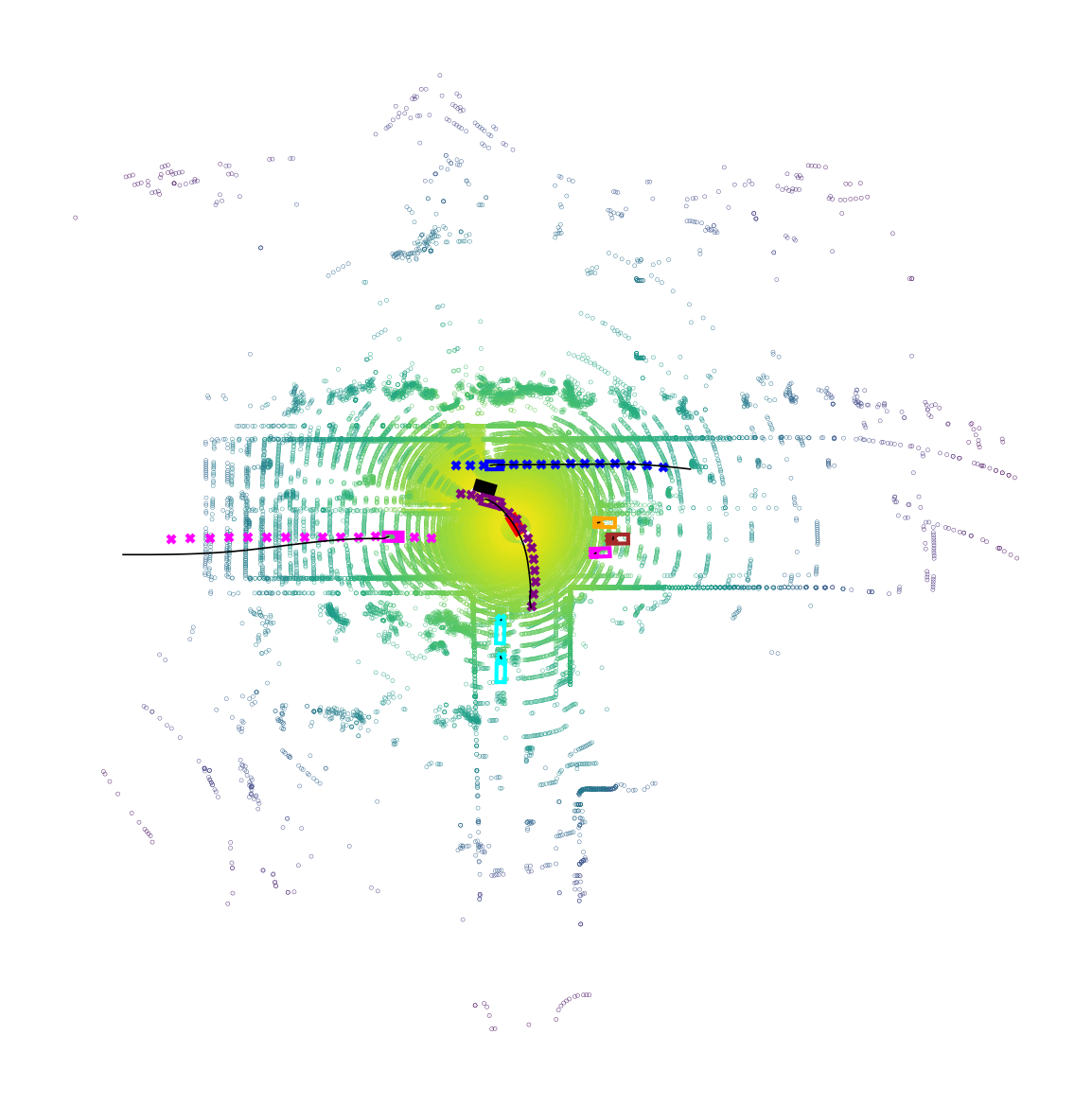}
        \caption{Cooperative Prediction (Ours)}
        \label{fig:case3y}
        
    \end{subfigure}
    
    \includegraphics[width=0.9\textwidth]{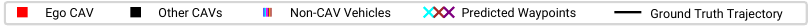}
    \vspace{-0.cm} 
    \caption{The visualizations of predicted trajectories under different model settings in two traffic scenarios. 
    In (a) and (c), some surrounding vehicles are not detected and the predicted trajectories (colored waypoints) without cooperation deviate significantly from the ground truth (black lines). In contrast, in (b) and (d) where cooperative prediction is enabled, the predicted trajectories become closer to the ground truth due to additional useful information from others.}
    \label{fig:traj_vis}
    \vspace{-0.4cm}
\end{figure*}

\begin{figure}[!]
    \centering
    \includegraphics[width=\columnwidth]{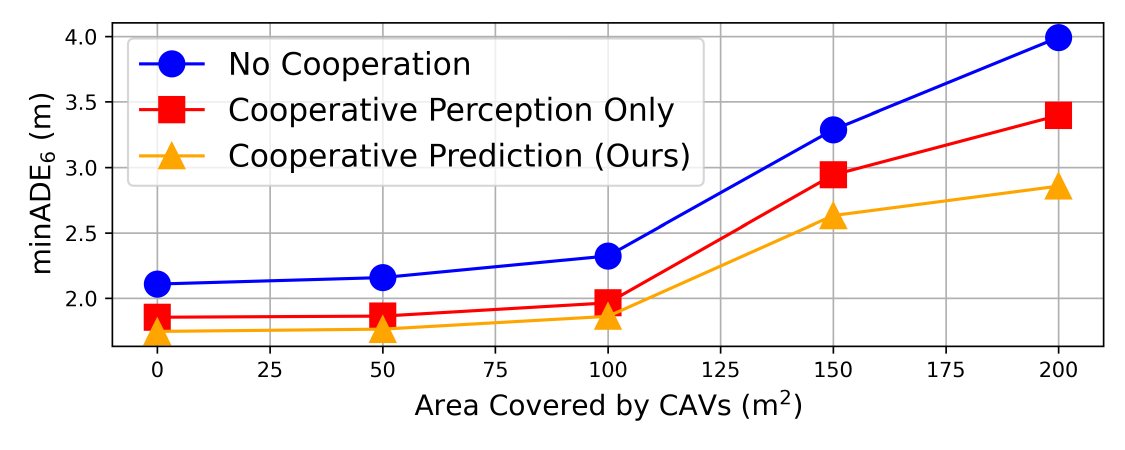}
    \vspace{-0.6cm} 
    \caption{A comparison of motion prediction performance at 5s prediction horizon under different areas covered by CAVs in OPV2V. The area is calculated based on the smallest convex hull that covers all the CAVs. As the number of CAVs increases in different scenarios, more areas are likely covered, which boosts the performance gap between no cooperation and cooperative prediction.
    }
    \vspace{-0.6cm}
    \label{fig:CAV_area}
\end{figure}

\vspace{-0.3cm}
\subsection{Quantitative and Qualitative Results with Ablation}
Since the end goal of our work is cooperative motion prediction, we first present and analyze the quantitative results of prediction followed by the object detection and tracking performance as intermediate steps. Additionally, we provide qualitative comparisons of the predicted trajectories between our framework and the no-cooperation setting.

\textbf{Cooperative Motion Prediction.}
We present a series of quantitative and ablation studies on cooperative motion prediction. The detailed results are shown in Table~\ref{tab:motion_prediction} and Fig. \ref{fig:CAV_area}.
CMP (Ours) with the \textit{Cooperative Perception} setting does not include our prediction aggregation module, and CAVs only share the compressed BEV features in the perception stage.
Table~\ref{tab:motion_prediction} shows that CMP (Ours) with the \textit{Cooperative P\&P} setting enhances the prediction performance by a large margin and the improvement becomes more significant as the prediction horizon increases. Also, it outperforms the cooperative perception and prediction network proposed by the strongest baseline V2VNet \cite{wang2020v2vnet}. 

Our cooperative motion prediction (CMP) framework with cooperation in both perception and prediction stages leads to even greater improvement. In OPV2V at 5s prediction horizon, CMP achieves a 12.3\%/16.4\% reduction in minADE$_6$ and a 15.1\%/19.7\% reduction in minFDE$_6$ compared with the \textit{V2VNet} and \textit{No Cooperation} settings, respectively. Similarly, in V2V4Real at the same prediction horizon, CMP achieves a 12.0\%/22.2\% reduction in minADE$_6$ and a 7.3\%/19.0\% reduction in minFDE$_6$.
The reason is that cooperative perception improves the detection accuracy and thus the quality of historical trajectories employed by the prediction module.
Moreover, the prediction aggregation module allows CAVs to leverage the predictions from others to collectively compensate for their prediction in challenging situations.
In Fig. \ref{fig:CAV_area}, our cooperation modules bring more benefits as the CAVs in the scene cover larger fields of view. Specifically, we approximate the perception coverage area from the CAVs as the area of the convex hull formed by the CAVs.
As this area grows to over \SI{200}{m^2}, the performance improvement of our model climbs to 17.5\%/28.4\% compared to the other settings.
The reason is that the communication between CAVs enhances the situational awareness of the ego car with more comprehensive, precise detections of surrounding objects and richer insights for future prediction.
Our bandwidth-efficient framework supports simultaneous transmission between up to ten vehicles under typical V2V network conditions \cite{wang2020v2vnet}. Moreover, our framework dynamically adjusts to changing network topology by aggregating predictions from active CAVs in each frame. This topology-agnostic design ensures robust performance even with varying connectivity or link duration, as evidenced by stable prediction accuracy across diverse traffic scenarios in OPV2V \cite{opv2v} and V2V4Real \cite{xu2023v2v4real}. 

\textbf{Qualitative Results.}
We provide visualizations of the predicted vehicle trajectories in two scenarios in Fig. ~\ref{fig:traj_vis} to show the effectiveness of cooperative prediction.
Fig. ~\ref{fig:traj_vis}(a) and ~\ref{fig:traj_vis}(b) depict the same scenario involving two CAVs. It shows that cooperative prediction significantly reduces the number of non-CAV vehicles that are overlooked, which highlights the enhanced sensing capability brought by cooperation, allowing each CAV to extend its perception range and detect vehicles that might otherwise be missed.
Fig. ~\ref{fig:traj_vis}(c) and ~\ref{fig:traj_vis}(d) show another scenario, demonstrating the improved accuracy of cooperative prediction. In this case, the predicted trajectories by cooperative prediction align more closely with the ground truth thanks to information sharing between CAVs.

\begin{table*}[!t]
\centering
\footnotesize
\setlength{\tabcolsep}{2pt}
\caption{The comparisons of cooperative object detection performance. Upper: OPV2V. Lower: V2V4Real.}
\label{tab:perception}
{
    \begin{tabular}{c|m{2cm}<{\centering}|m{1.5cm}<{\centering}|c|c|c|c|c|c|c|c|c|m{1.5cm}<{\centering}}
    \toprule
    \midrule
     Model & Communication Setting & Compression Ratio & AP 0.3$\uparrow$ & AR 0.3$\uparrow$ & F1 0.3$\uparrow$ & AP 0.5$\uparrow$ & AR 0.5$\uparrow$ & F1 0.5$\uparrow$ &AP 0.7$\uparrow$ & AR 0.7$\uparrow$ & F1 0.7$\uparrow$ & Bandwidth (MB/s)$\downarrow$ \\
    \midrule
    SinBEVT \cite{xu2022cobevt} & No Cooperation                    & N/A  &0.80&0.43&0.56&0.79&0.43&0.55&0.65&0.39&0.48& N/A \\
    \cmidrule{1-13}
    V2VNet \cite{wang2020v2vnet}
    & \SI{100}{ms} Delay& None&0.86&\textbf{0.49}&0.63&0.83&0.48&0.61&0.66&0.43&0.52&82.5\\
    \cmidrule{1-13} 
    \multirow{2}{*}{CoBEVT \cite{xu2022cobevt}} 
    &\SI{100}{ms} Delay&None   &\textbf{0.94}&\textbf{0.49}&\textbf{0.65}&\textbf{0.93}&\textbf{0.49}&\textbf{0.64}&0.81&\textbf{0.45}&\textbf{0.58}&82.5\\
    &\SI{100}{ms} Delay&256$\times$ &0.93&0.47&0.63&0.92&0.47&0.62&\textbf{0.82}&0.44&\textbf{0.58}&0.32\\
    \bottomrule
    \end{tabular}
}

\vspace{0.2cm}
{
    \begin{tabular}{c|m{2cm}<{\centering}|m{1.5cm}<{\centering}|c|c|c|c|c|c|c|c|c|m{1.5cm}<{\centering}}
    \toprule
    \midrule
     Model & Communication Setting & Compression Ratio & AP 0.3$\uparrow$ & AR 0.3$\uparrow$ & F1 0.3$\uparrow$ & AP 0.5$\uparrow$ & AR 0.5$\uparrow$ & F1 0.5$\uparrow$ &AP 0.7$\uparrow$ & AR 0.7$\uparrow$ & F1 0.7$\uparrow$ & Bandwidth (MB/s)$\downarrow$ \\
    \midrule
    SinBEVT \cite{xu2022cobevt} & No Cooperation                    & N/A  &0.52&0.33&0.40&0.42&0.29&0.35&0.19&0.20&0.20& N/A \\
    \cmidrule{1-13} 
    V2VNet \cite{wang2020v2vnet} & \SI{100}{ms} Delay&None&0.70&\textbf{0.41}&0.52&0.57&0.36&0.44&0.30&0.26&0.28&60.0\\  
    \cmidrule{1-13} 
    \multirow{2}{*}{CoBEVT \cite{xu2022cobevt}} 
    &\SI{100}{ms} Delay&None&\textbf{0.72}&\textbf{0.41}&\textbf{0.53}&0.61&\textbf{0.37}&\textbf{0.46}&\textbf{0.32}&\textbf{0.27}&\textbf{0.29}&60.0\\
    & \SI{100}{ms} Delay&256$\times$&\textbf{0.72}&\textbf{0.41}&0.52&\textbf{0.62}&\textbf{0.37}&\textbf{0.46}&0.31&0.26&0.28&0.23\\
    \bottomrule
    \end{tabular}
    \vspace{-0.2cm}
    }
\end{table*}

\begin{table*}[!t]
\centering
\footnotesize
\setlength{\tabcolsep}{2pt}
\caption{The comparisons of multi-object tracking performance. Upper: OPV2V. Lower: V2V4Real.}
\label{tab:tracker}
\label{trackinig}
{
    \begin{tabular}{c|m{2cm}<{\centering}|m{2cm}<{\centering}|m{1.4cm}<{\centering}|m{1.4cm}<{\centering}|m{1.4cm}<{\centering}|m{1.4cm}<{\centering}|m{1.4cm}<{\centering}|m{1.4cm}<{\centering}|m{1.4cm}<{\centering}}
    \toprule
    \midrule
    Model & Communication Setting & Compression Ratio   & sAMOTA $\uparrow$ & AMOTA $\uparrow$ & AMOTP $\uparrow$ & MOTA $\uparrow$ & MOTP $\uparrow$ & MT $\uparrow$ & ML $\downarrow$ \\
    \midrule
    SinBEVT \cite{xu2022cobevt} & No Cooperation & N/A 
    & 44.99 & 20.19 & 21.44 & 44.87 & 47.64 & 21.23 & 34.46\\
    \cmidrule{1-10} 
    \multirow{2}{*}{CoBEVT \cite{xu2022cobevt}} & \multirow{2}{*}{$\SI{100}{ms}$ Delay} & None 
    & \textbf{62.48}& \textbf{38.76} & 29.93 & \textbf{62.02} & 47.89 & \textbf{38.77} & \textbf{16.62}\\
    && 256$\times$
    & 62.43 & 37.98  & \textbf{30.20} & 60.76 & \textbf{48.31} & 36.92 & 17.85 \\
    \bottomrule
    \end{tabular}
}

\vspace{0.2cm}
{
    \begin{tabular}{c|m{2cm}<{\centering}|m{2cm}<{\centering}|m{1.4cm}<{\centering}|m{1.4cm}<{\centering}|m{1.4cm}<{\centering}|m{1.4cm}<{\centering}|m{1.4cm}<{\centering}|m{1.4cm}<{\centering}|m{1.4cm}<{\centering}}
    \toprule
    \midrule
    Model & Communication Setting & Compression Ratio   & sAMOTA $\uparrow$ & AMOTA $\uparrow$ & AMOTP $\uparrow$ & MOTA $\uparrow$ & MOTP $\uparrow$ & MT $\uparrow$ & ML $\downarrow$ \\
    \midrule
    SinBEVT \cite{xu2022cobevt} & No Cooperation & N/A 
    & 32.43 & 10.35 & 21.62 & 31.66 & 66.15 & 9.94 & 68.51 \\
    \cmidrule{1-10}
    \multirow{2}{*}{CoBEVT \cite{xu2022cobevt}} & \multirow{2}{*}{$\SI{100}{ms}$ Delay} & None & \textbf{44.88}& \textbf{19.31} & \textbf{31.12} & \textbf{42.77} & \textbf{69.00} & \textbf{12.91} & 62.59\\
    && 256$\times$ 
    & 41.14 & 16.27 & 28.21 & 39.40 & 68.31 & 10.66 & \textbf{62.24}\\
    \bottomrule
    \end{tabular}
}
\vspace{-0.3cm}
\end{table*}

\textbf{Cooperative Object Detection.} 
In Table \ref{tab:perception}, we demonstrate the effects of multi-vehicle cooperation, communication delay, and compression ratio of BEV features on the object detection performance. 
The comparisons between \textit{No Cooperation} and other settings show the improvement brought by the CAV communications.
We evaluated compression ratios of 64$\times$, 128$\times$, and 256$\times$ to study trade-offs between bandwidth usage and model performance. Results indicate that the 256$\times$ compression ratio achieves a significant bandwidth reduction from \SI{82.5}{MB/s} to \SI{0.32}{MB/s} in OPV2V, with only a 4\% decrease in detection accuracy, validating its feasibility and suitability for constrained V2V communication environments.
The compression only takes \SI{0.5}{ms}, making its impact negligible compared with the 100 ms transmission delay, thus demonstrating efficiency in data transmission when sharing BEV features from one CAV to another. 
Results show that CoBEVT outperforms V2VNet on OPV2V and achieves comparable detection performance on V2V4Real.
Based on these findings, we adopt a 256$\times$ compression ratio for BEV features and accommodate a $\SI{100}{ms}$ communication latency between CAVs, which balances between model performance and real-world hardware constraints (i.e., bandwidth, latency).

\textbf{Tracking.} 
We show the enhancement of tracking performance enabled by multi-vehicle cooperations in Table~\ref{tab:tracker}. 
V2V communication enables the fusion of perception information across different CAVs, which significantly increases the number of true positives (i.e., accurately detected objects) and reduces the instances of false positives and false negatives (i.e., missing objects). We observe a higher MOTA and MOTP in OPV2V compared to V2V4Real. The improvement in object detection is a major cause for the enhanced performance of the tracking system.  Furthermore, despite the substantial BEV feature compression, we observe no detrimental effect on the tracking performance in both OPV2V and V2V4Real, which implies that tracking remains robust even under significant feature compression.

\vspace{-0.3cm}
\subsection{Real-time Performance Analysis}
\begin{figure}[!]
    \centering
    \includegraphics[width=\columnwidth]{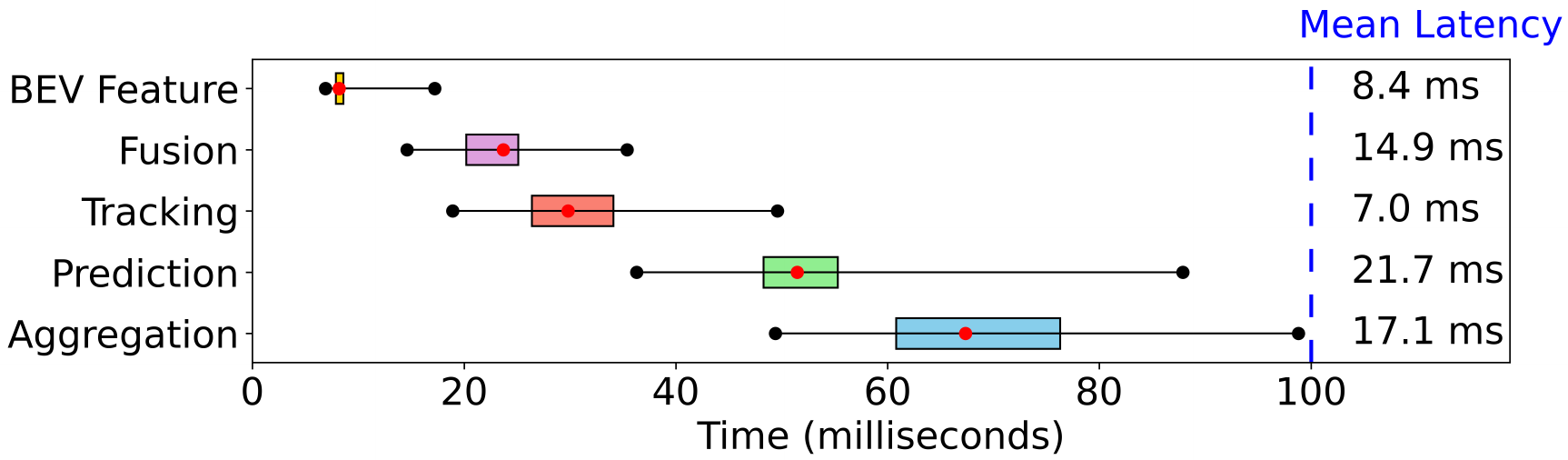}
    \caption{
    Micro-benchmarking of the pipeline latency across different modules. Each bar represents the cumulative latency distribution for a module. Colored rectangles show the 25th to 75th percentile range, whiskers denote minimum and maximum values, and red dots indicate medians. The average latency for each module is displayed on the right.}
    \vspace{-0.3cm}
    \label{fig:micro}
\end{figure}

In Fig. \ref{fig:micro}, micro-benchmarking of our pipeline on a single NVIDIA 6000 Ada GPU shows an average runtime of 67.3 ms, well below the 100 ms communication delay constraint, making it feasible for real-time onboard vehicle deployment.

\section{Conclusion}

In this paper, we introduce the first-of-its-kind cooperative motion prediction framework that advances the cooperative capabilities of CAVs, addressing the crucial need for safe and robust decision making in dynamic environments. 
By integrating cooperative perception with trajectory prediction, our work marks a pioneering effort in the realm of connected and automated vehicles, which enables CAVs to share and fuse data from LiDAR point clouds to improve object detection, tracking, and motion prediction. 
Specifically, our contributions include a latency-robust cooperative prediction pipeline, communication bandwidth analysis, and a cooperative aggregation mechanism for motion prediction, which advance CAV performance and set a benchmark for future research.
In this work, our framework does not yield a fully end-to-end approach due to the non-differentiable tracker. 
Future work will develop a fully differentiable end-to-end pipeline that jointly optimizes perception, prediction, and aggregation to minimize accumulative errors and cross-module inconsistencies.
We will also investigate multi-modal sensor fusion with heterogeneous CAVs to improve flexibility.

\printbibliography
\end{document}